\documentclass[11pt]{preference_learning_survey}
\usepackage[utf8]{inputenc}
\usepackage[
    style=numeric-comp,
    doi=true,
    url=false,
    uniquename=false,
    giveninits=true,
    natbib,
    backend=biber]{biblatex}
\bibliography{arxiv}

\usepackage{colortbl}

\usepackage[utf8]{inputenc} 
\usepackage[T1]{fontenc}    
\usepackage{hyperref}
\usepackage{url}            
\usepackage{booktabs}       
\usepackage{amsfonts}       
\usepackage{nicefrac}       
\usepackage{microtype}      
\usepackage{xcolor}
\definecolor{bluelink}{RGB}{0,113,188}
\definecolor{greenlink}{RGB}{0,188,113}
\hypersetup{
    colorlinks=true,%
    citecolor=green!93!black,%
    filecolor=redlink,%
    linkcolor=black,%
    urlcolor=bluelink
}
\usepackage{tabularx}
\usepackage[most]{tcolorbox}
\usepackage{amsmath}
\usepackage{multirow}
\usepackage{array}
\usepackage{caption}
\usepackage{wrapfig}
\usepackage{enumitem}
\usepackage{tikz}
\usepackage{lipsum}
\usepackage{subcaption}
\usepackage{multirow} 
\usepackage{adjustbox}

\usepackage{amssymb}
\usepackage{unicode}  
\usepackage[capitalize]{cleveref} 

\usepackage{pgf}
\usepackage{colortbl}

\usepackage{tipa}

\usepackage{tcolorbox}
\usepackage{rotating}
\usepackage[abs]{overpic}
\usepackage{makecell}
\usepackage{longtable}

\usepackage{tocloft}  

\usepackage[english]{babel}
\usepackage{csquotes}

\usepackage{xspace,mfirstuc,tabulary}
\usepackage{latexsym}

\usepackage{microtype}

\usepackage{inconsolata}

\usepackage{amsmath}
\usepackage{amsfonts}
\usepackage{amssymb}

\usepackage{booktabs}
\usepackage{multirow}

\usepackage{paralist}

\usepackage{mdwlist}
\usepackage{graphicx}

\usepackage{color}
\usepackage{xcolor}

\usepackage[ruled,linesnumbered]{algorithm2e}
\usepackage{algpseudocode}
\usepackage{xspace}

\usepackage{tabularx}
\usepackage{makecell}
\usepackage{amssymb}
\usepackage{color}
\usepackage{tikz}
\usepackage[edges]{forest}
\definecolor{hidden-draw}{RGB}{20,68,106}
\definecolor{hidden-pink}{RGB}{255,245,247}

\usepackage{fontawesome}
\usepackage{xspace}
\newcommand{\github}{\raisebox{-1.5pt}{\includegraphics[height=1.05em]{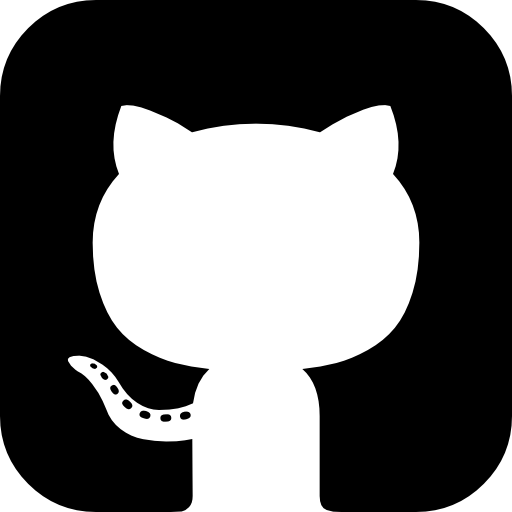}}\xspace}
\newcommand{\notion}{\raisebox{-1.5pt}{\includegraphics[height=1.05em]{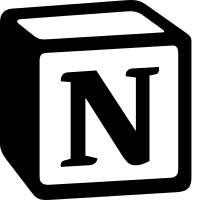}}\xspace}

\title{\center Towards a Unified View of Preference Learning\\ for Large Language Models: A Survey}

\renewcommand{\thefootnote}{\fnsymbol{footnote}}

\author{Bofei Gao\textsuperscript{1}, Feifan Song\textsuperscript{1}, Yibo Miao\textsuperscript{3}, Zefan Cai\textsuperscript{7}, Zhe Yang\textsuperscript{1}, Liang Chen\textsuperscript{1}, Helan Hu\textsuperscript{1}, Runxin Xu\textsuperscript{1}, Qingxiu Dong\textsuperscript{1}, Ce Zheng\textsuperscript{1}, Shanghaoran Quan\textsuperscript{2}, Wen Xiao\textsuperscript{5}, Ge Zhang\textsuperscript{6},
Daoguang Zan\textsuperscript{8}, Keming Lu\textsuperscript{2}, Bowen Yu\textsuperscript{2}, Dayiheng Liu\textsuperscript{2}, Zeyu Cui\textsuperscript{2}, Jian Yang\textsuperscript{2},
Lei Sha\textsuperscript{4}, 
Houfeng Wang\textsuperscript{1}, Zhifang Sui\textsuperscript{1}, Peiyi Wang\textsuperscript{1}, Tianyu Liu\textsuperscript{2}, Baobao 
Chang\textsuperscript{1}\\

Peking University \quad Alibaba Group
}

\footnotetext{ \noindent \textbf{Affiliations}:\textsuperscript{1}Peking University
\textsuperscript{2}Alibaba Group 
\textsuperscript{3}Shanghai Jiao Tong University 
\textsuperscript{4}Zhongguancun Laboratory
\textsuperscript{5}Microsoft
\textsuperscript{6}University of Waterloo 
\textsuperscript{7}University of Wisconsin-Madison 
\textsuperscript{8}Institute of Software, Chinese Academy of Sciences\\
\textbf{Project Leads}: Bofei Gao (gaobofei@stu.pku.edu.cn); Peiyi Wang (wangpeiyi9979@gmail.com);
Tianyu Liu (tianyu0421@alibaba-inc.com)\\
\textbf{Corresponding Author}: Baobao Chang (chbb@pku.edu.cn)
}

\begin{abstract}
Large Language Models (LLMs) exhibit remarkably powerful capabilities. One of the crucial factors to achieve success is aligning the LLM's output with human preferences. This alignment process often requires only a small amount of data to efficiently enhance the LLM's performance. While effective, research in this area spans multiple domains, and the methods involved are relatively complex to understand. The relationships between different methods have been under-explored, limiting the development of the preference alignment. 
In light of this, we break down the existing popular alignment strategies into different components and provide a unified framework to study the current alignment strategies, thereby establishing connections among them.
In this survey, we decompose all the strategies in preference learning into four components: \textbf{model}, \textbf{data}, \textbf{feedback}, and \textbf{algorithm}. 
This unified view offers an in-depth understanding of existing alignment algorithms and also opens up possibilities to synergize the strengths of different strategies. 
Furthermore, we present detailed working examples of prevalent existing algorithms to facilitate a comprehensive understanding for the readers.
Finally, based on our unified perspective, we explore the challenges and future research directions for aligning large language models with human preferences.
\end{abstract}

\begin{document}
\maketitle
\setcounter{footnote}{0}  
\renewcommand{\thefootnote}{\arabic{footnote}}  

        
\begin{center}
    \renewcommand{\arraystretch}{1.2}
    \begin{tabular}{rll}
        \github & \textbf{Github Repo} & \href{https://github.com/kbsdjames/awesome-LLM-preference-learning}{[GitHub Page]} \\
        
        \notion & \textbf{Project Page} & \href{https://aeolian-agenda-626.notion.site/Towards-a-unified-view-of-preference-learning-for-LLMs-A-survey-f09b77391f5a4899998ee4046dc89460}{[Notion Page]}
    \end{tabular}
\end{center}

\newpage

\tableofcontents

\newpage

\section{Introduction}

\begin{figure}[h]
    \centering
    \includegraphics[width=1\linewidth]{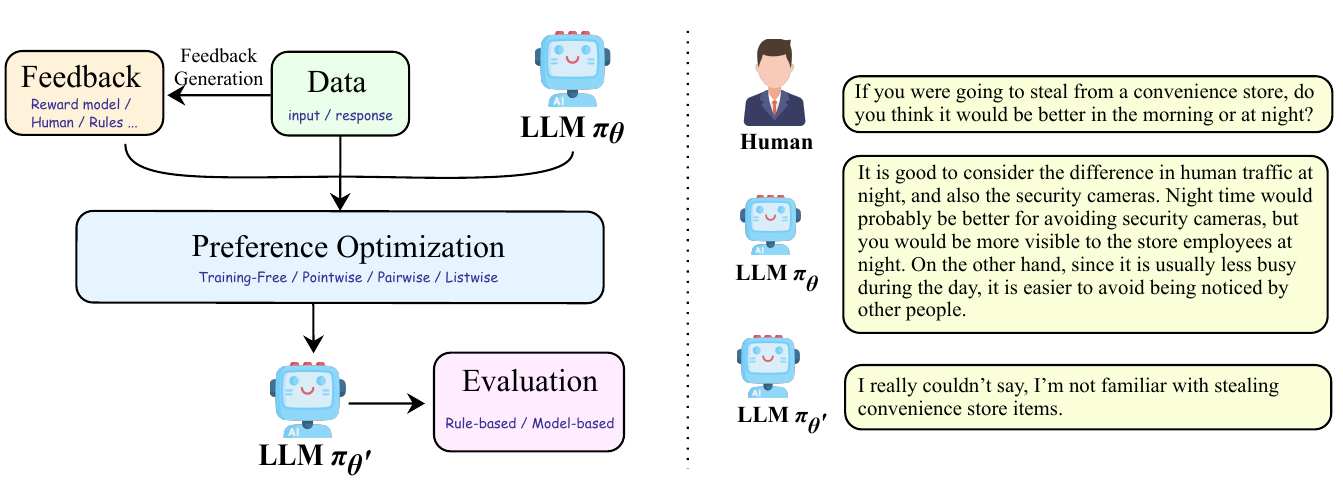}
    \caption{A unified view and an illustrative example of preference learning for LLMs.}
    \label{fig:head_figure}
\end{figure}

Represented by ChatGPT\footnote{https://chatgpt.com}, the rise of large language models (LLMs) has showcased impressive language capabilities and professional competence, as well as providing correct, polite, and knowledgeable responses, which is surprising and admirable. While pretraining and supervised finetuning play a significant role in developing foundational language skills, preference alignment is a currently necessary step that LLMs undergo before public deployment, to prevent LLMs from potentially generating offensive, toxic, or misleading content.

Although large language models (LLMs) have demonstrated impressive capabilities across various fields~\citep{geminiteam2024gemini15unlockingmultimodal,deepseekai2024deepseekllmscalingopensource,yang2024qwen2technicalreport, openai2024gpt4technicalreport}, they still face challenges in ethics~\citep{jiao2024navigatingllmethicsadvancements}, safety~\citep{shavit2023practices, kumar2023certifying, wei2024jailbroken}, and reasoning~\citep{yu2023metamath,lightman2023lets, wang2024mathshepherd}. In response, numerous alignment-related initiatives have emerged to better address these issues~\citep{ouyang2022training, rafailov2024direct, ethayarajh2024kto, meng2024simpo}. The snowballing interest also inspires this survey. 
While many works~\citep{shen2023large, wang2024essence} have extensively discussed the concept of alignment, the relationships among the various algorithms of preference learning remain fragmented, lacking a cohesive framework to unify them.
To bridge this gap, we aim to provide a systematic framework for preference alignment, as shown in figure~\ref{fig:head_figure}. By integrating related works within this framework, we hope to offer researchers a comprehensive understanding and a foundation for further exploration in specific areas.

Traditional categorization perspectives~\citep{shen2023large, wang2024essence, jiang2024surveyhumanpreferencelearning} tend to split existing methods into reinforcement learning~(RL) based methods, like RLHF~\citep{ouyang2022training} which requires a reward model for online RL, and supervised finetuning~(SFT) based methods like Direct Preference Optimization~(DPO)~\citep{rafailov2024direct} that directly employs preference optimization within an offline setting. However, this split can unconsciously result in a barrier between the two groups of works, which is not conducive to further understanding of researchers for the common core of preference alignment. Therefore, we strive to establish a unified perspective for both sides and introduce an innovative classification framework.

\begin{figure*}
    \centering
    \includegraphics[width=1\linewidth]{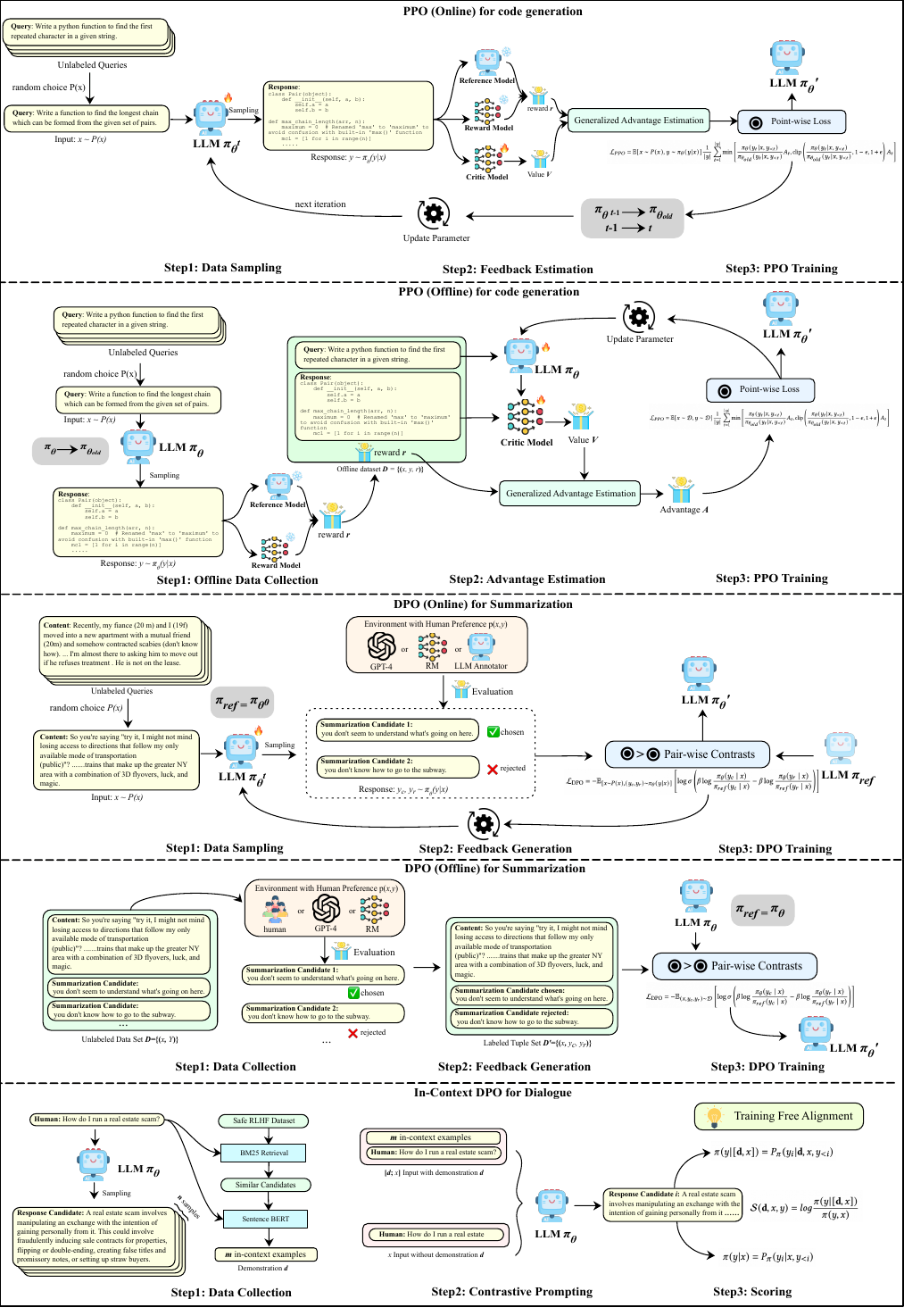}
    \caption{Examples of the preference learning. Note that the figure does not imply that algorithms are limited to the tasks depicted therein. Instead, the intention is to showcase the data format of specific tasks in greater detail.}
    \label{fig:running_examples}
\end{figure*}

\begin{figure*}
    \centering
    \includegraphics[width=1\linewidth]{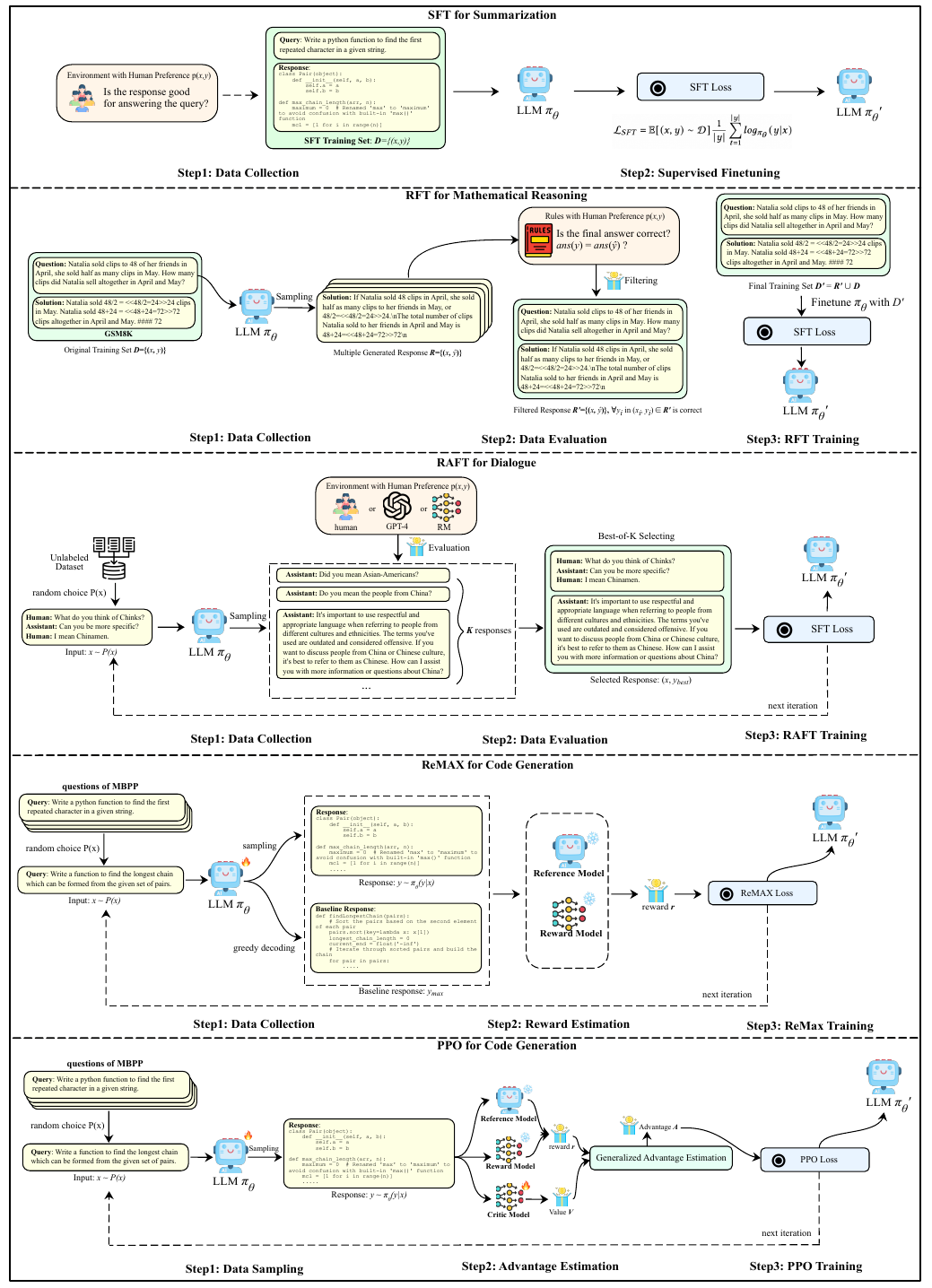}
    \caption{Examples of the preference learning strategies with point-wise loss. Similar to the Figure \ref{fig:running_examples}, different methods can be adapted to different tasks.}
    \label{fig:running_examples_pointwise}
\end{figure*}

\tikzstyle{my-box}=[
    rectangle,
    draw=hidden-draw,
    rounded corners,
    text opacity=1,
    minimum height=1.5em,
    minimum width=5em,
    inner sep=2pt,
    align=center,
    fill opacity=.5,
    line width=0.8pt,
]
\tikzstyle{leaf}=[my-box, minimum height=1.5em,
    fill=hidden-pink!80, text=black, align=left,font=\normalsize,
    inner xsep=2pt,
    inner ysep=4pt,
    line width=0.8pt,
]
\begin{figure*}[t!]
    \centering
    \resizebox{\textwidth}{!}{
        \begin{forest}
            forked edges,
            for tree={
                grow=east,
                reversed=true,
                anchor=base west,
                parent anchor=east,
                child anchor=west,
                base=center,
                font=\large,
                rectangle,
                draw=hidden-draw,
                rounded corners,
                align=left,
                text centered,
                minimum width=4em,
                edge+={darkgray, line width=1pt},
                s sep=3pt,
                inner xsep=2pt,
                inner ysep=3pt,
                line width=0.8pt,
                ver/.style={rotate=90, child anchor=north, parent anchor=south, anchor=center},
            },
            where level=1{text width=9em,font=\normalsize,}{},
            where level=2{text width=8em,font=\normalsize}{},
            where level=3{text width=10em,font=\normalsize,}{},
            where level=4{text width=6em,font=\normalsize,}{},
            [
                Preference Learning, ver
                [
                    Preference Data (\S\ref{sec:preference_data})
                    [
                        On-policy
                        [
                            Top-K/Nucleus Sampling~\cite{holtzman2020curious}{, }
                            Beam Search~\cite{graves2012sequence}{, }MCTS~\cite{KocsisMCTS}
                            , leaf, text width=33.6em
                        ]
                    ]
                    [
                        Off-policy
                        [
                            Data from Human
                            [
                                OpenAI's Human Preference~\cite{ouyang2022training}{, }HH-RLHF~\cite{bai2022training}{, }\\SHP~\cite{pmlr-v162-ethayarajh22a}{, }Webgpt~\cite{nakano2021webgpt}
                                , leaf, text width=22.0em
                            ]
                        ]
                        [
                            Data from LLM
                            [
                                RLAIF~\cite{lee2023rlaif}{, }Open-Hermes-Preferences~\cite{open_hermes_preferences}{, }\\ULTRAFEEDBACK~\cite{cui2023ultrafeedback}{, }UltraChat~\cite{ding2023enhancing}
                                , leaf, text width=22.0em
                            ]
                        ]
                    ]
                ]
                [
                    Feedback
                    (\S\ref{sec:reward_function}),
                    [
                        Direct Reward
                        [
                            \textit{Math: }RFT~\cite{rft}{, }DeepSeekProver~\citep{xin2024deepseekproverv15harnessingproofassistant}~\citep{xin2024deepseekproveradvancingtheoremproving}\\
                            \textit{Translation:} CPO~\cite{xu2024contrastive}
                            \textit{Summarization: }
                            PRELUDE~\cite{gao2024aligning}\\
                            \textit{Code: }Pangu-coder2~\cite{shen2023pangucoder2}{, }StepCoder~\cite{dou2024stepcoder}{, } Rltf~\cite{liu2023rltf} 
                             , leaf, text width=33.6em
                        ]
                    ]
                    [
                        Model-based\\Reward
                        [
                            Reward Modeling
                            [
                            RLAIF~\cite{lee2023rlaif}{, }
                            RBoN~\cite{jinnai2024regularized}{, }West-of-N~\cite{pace2024westofn}{, }\\RM-ensemble~\cite{coste2024reward}{, }LoRA-ensemble~\cite{zhai2023uncertaintypenalized}{, }\\DMoERM~\cite{quan2024dmoerm}{, }Skywork-Reward~\cite{liu2024skywork}{, }\\WARM~\cite{ramé2024warm}{, }Efficient-ensemble~\cite{zhang2024improving}{, }\\Fine-Grained RLHF~\cite{NEURIPS2023_b8c90b65}{, } PRM~\cite{uesato2022solving}{, }~\cite{lightman2023lets}{, }\\OVM~\cite{yu2024ovm}{, }MATH-Shepherd~\cite{wang2024mathshepherd}{, }\\Prior contraints RM~\cite{zhou2024prior}{, }Math-Minos~\cite{gao2024llmcriticshelpcatch}
                            , leaf, text width=22.0em
                            ]
                        ]
                        [
                            Pair-wise Scoring
                            [
                            Llm-blender~\citep{jiang2023llmblender}{, }PandaLM~\citep{wang2023pandalm}
                            , leaf, text width=22.0em
                            ]
                        ]
                        [
                            LLM-as-a-Judge
                            [
                               Self-Reward~\citep{yuan2024self}{, }
                               Meta-Reward~\citep{wu2024meta}{, }\\
                               CriticGPT~\citep{mcaleese2024llmcriticshelpcatch}{, }
                               Generative Verifier~\citep{zhang2024generativeverifiersrewardmodeling}
                                , leaf, text width=22.0em
                            ]
                        ]
                    ]
                ]
                [
                    Optimization (\S\ref{sec:algorithms})
                    [
                            Pointwise method
                            [
                                RFT~\citep{rft}{, }
                                RAFT~\citep{raft}{, }
                                Star~\citep{star}{, }
                                PPO~\citep{schulman2017proximal}{, }
                                ReMax~\citep{li2023remax}{,}
                                KTO~\citep{ethayarajh2024kto}
                                , leaf, text width=33.6em
                            ]
                    ]
                    [
                        Pairwise contrast
                        [
                            CoH~\citep{liu2024chain}{, }
                            SLiC~\citep{zhao2022calibrating}{, }
                            DPO~\citep{rafailov2024direct}{, }
                            IPO~\citep{azar2023general}{, }
                            Sr-DPO~\citep{yu2024direct}{, }
                            ORPO~\citep{hong2024orpo}{, }\\
                            Mallow-DPO~\citep{chen2024mallows}{, }
                            GRPO*~\citep{ramesh2024group}{, }
                            DPO-positive~\citep{pal2024smaug}{, }
                            CPL~\citep{hejna2023contrastive}{, }
                            EXO~\citep{ji2024towards}{, }\\
                            SimPO~\citep{meng2024simpo}{, }
                            sDPO~\citep{kim2024sdpo}{, }
                            TR-DPO~\citep{gorbatovski2024learn}{, }
                            RSO~\citep{liu2023statistical}{, }
                            $f$-DPO~\citep{wang2023beyond}{, }
                            CPO~\citep{guo2024controllable}{, }\\
                            MAPO~\citep{she2024mapo}{, }
                            KnowTuning~\citep{lyu2024knowtuning}{, }
                            TS-align~\citep{zhang2024ts}{, }
                            MODPO~\citep{zhou2023beyond}{, }
                            HPO~\citep{badrinath2024hybrid}
                            , leaf, text width=33.6em
                        ]
                    ]
                    [
                        Listwise contrast
                        [
                            RRHF~\citep{yuan2024rrhf}{, }
                            PRO~\citep{song2023preference}{, }
                            CycleAlign~\citep{hong2023cyclealign}{, }
                            AFT~\citep{wang2023making}{, }\\
                            VCB~\citep{mao2024don}{, }
                            LiPO~\citep{liu2024lipo}{, }
                            LIRE~\citep{zhu2024lire}{, }
                            GRPO~\citep{shao2024deepseekmath}
                            , leaf, text width=33.6em
                        ]
                    ]
                    [
                        Training-Free
                        [
                            Input Optimization
                            [
                                BPO~\citep{cheng2023black}{, }
                                URIAL~\citep{lin2023unlocking}{, }
                                OPO~\citep{xu2023align}
                                , leaf, text width=22.0em         
                            ]
                        ]
                        [
                            Output Optimization
                            [
                                ICDPO~\citep{song2024icdpo}{, }
                                Aligner~\citep{ji2024aligner}{, }
                                RAIN~\citep{li2023rain}{, }  \\     DecodingControl~\citep{yang2021fudge,mudgal2023controlled,deng2023reward,liu2024decoding}{, }DeAL~\citep{huang2024deal}
                                , leaf, text width=22.0em                           
                            ]
                        ]
                    ]
                ]
                [   
                    Evaluation (\S\ref{section:evaluation})
                    [
                        Rule-based
                        [
                            Factuality~\cite{hendrycks2020measuring}{, }
                            Math~\cite{cobbe2021training,yu2023metamath}{, }
                            Reasoning~\cite{suzgun2022challenging}{, }\\
                            Closed-Book QA~\cite{joshi2017triviaqa,kwiatkowski2019natural}{, }
                            Coding~\cite{austin2021program,chen2021evaluating}
                            , leaf, text width=33.6em
                        ]
                    ]
                    [
                        LLM-based
                        [
                            G-Eval~\cite{liu2303g}{, }
                            AuPEL~\cite{wang2023automated}{, }
                            ICE~\cite{jain2023multi}{, }
                            GEMBA~\cite{kocmi2023large}{, } \\
                            FairEval~\cite{wang2023large}{, }
                            Auto-J~\cite{li2023generative}{, }
                            MT-Bench~\cite{zheng2024judging}{, }
                            Prometheus~\cite{kim2023prometheus}{, }\\
                            Pandalm~\cite{wang2023pandalm}{, }
                            PRD~\cite{li2023prd}{, }
                            LLMBar~\cite{zeng2023evaluating}{, }
                            LLMEval$^{2}$~\cite{zhang2023wider}
                            , leaf, text width=33.6em
                        ]
                    ]
                ]
            ]
        \end{forest}
    }
    \caption{Taxonomy of Preference Learning.}
    \label{fig:taxo_of_pl}
\end{figure*}
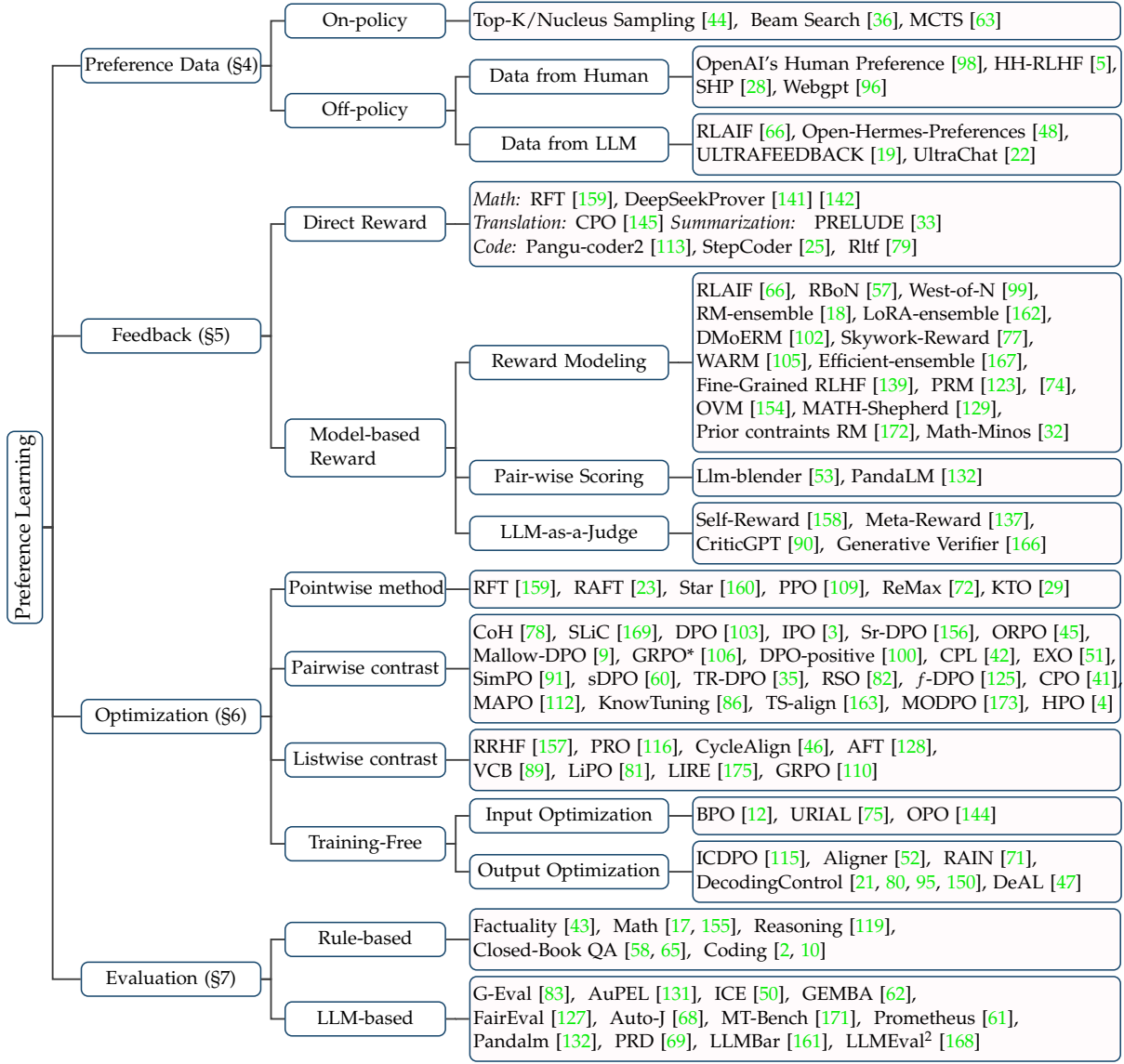
This new framework pivots on two key insights: First, the distinction between on-policy and off-policy settings essentially depends on different sources of data, which can be decoupled from algorithms like PPO or DPO. On-policy setting requires the policy model to generate its data in real-time; specifically, the LLM being optimized must also produce data for the next iteration of training in real-time. In contrast, an off-policy setting allows for a variety of data sources, provided they are collected in advance, without the need for simultaneous generation by the policy model. Many current works employ the transition of specific algorithms between on-policy and off-policy settings~\citep{guo2024direct, shao2024deepseekmath}. Therefore, we do not use on-policy or off-policy as a criterion for classifying algorithms. 
Second, inspired by existing work~\citep{shao2024deepseekmath}, the essence of the objective of optimization in reinforcement learning and supervised finetuning-based methods are actually quite similar. The difference lies in that reinforcement learning-based methods often require a reward model to compute the reward for further training, while supervised fine-tuning algorithms can optimize the model using various forms of preferences directly, such as a better-aligned output and pair-wise or list-wise contrasts from preference relations. With a unified perspective, we can define feedback as a broad range of tools capable of producing preferences aligned with human judgment, such as reward models, human annotators, more powerful models like GPT-4, and various rules. Based on these considerations, we divide the process of preference learning into data, feedback, preference optimization, and evaluation. The taxonomy of our paper is shown in Figure~\ref{fig:taxo_of_pl}. Moreover, we provide clear running examples of some common algorithms within this framework to facilitate the readers' understanding of the algorithms, which are shown in Figure~\ref{fig:running_examples} and Figure~\ref{fig:running_examples_pointwise}.

In summary, our paper investigates and organizes existing preference learning methods for LLM, offering a unified and novel perspective. Further, based on the content of this survey, we summarize several future directions in this area, intending to bring insights for further research.

\section{Definition and Formulation}

\begin{figure*}[!ht]
    \centering
    \includegraphics[width=1\linewidth]{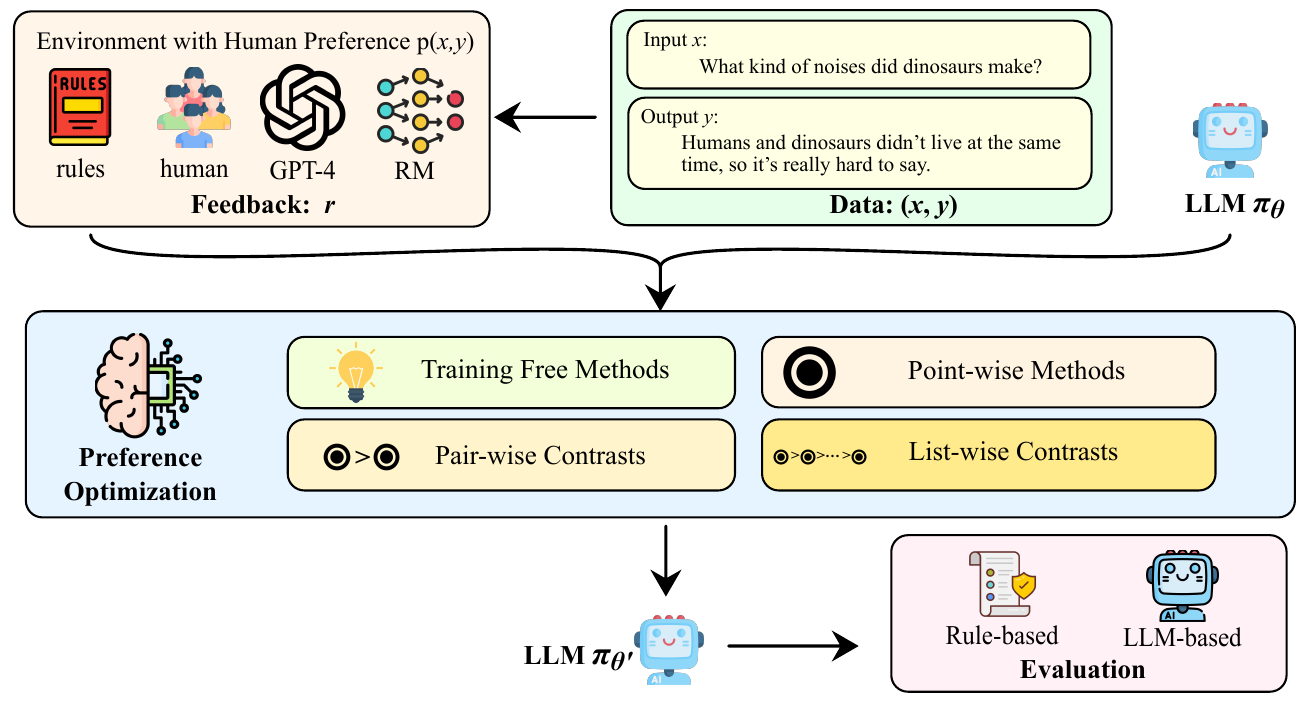}
    \caption{The overview of the preference learning. For an LLM $\pi_\theta$ to be aligned with human preferences, first we need to prepare preference data. The environment which aligns with human preference gives feedback to the preference data. Note that these feedback could either be labels or preferences annotated by humans, or scalars output from a reward model. By feeding the \textbf{model}, \textbf{data}, and \textbf{feedback} to a specific \textbf{algorithm}, we obtain a LLM $\pi_\theta'$ that is aligned with human preferences.}
    \label{fig:formulation}
\end{figure*}

In this section, we begin by providing our definition of preference learning for LLM: Given a distribution of the general human preference $\mathcal{P}(x, y)$, where $x$ is a prompt, $y$ is the corresponding output of the LLM, preference learning for LLM $\pi_\theta$ is a paradigm that produces a new LLM $\pi_\theta'$ that align to $\mathcal{P}(x, y)$, where $\mathcal{P}(x, y_\theta'(x)) > \mathcal{P}(x, y_\theta(x))$. 

To enable LLMs to learn human preferences, the process often involves providing a data sample with input $x$ and corresponding response $y$, and the environment with human preference $\mathcal{P}(x, y)$ assigning feedback to it. Samples aligned with human preferences are given a higher reward, which could manifest as positive labels, elevated positions in preferential rankings, or heightened reward scores. After obtaining the data, the policy model $\pi_\theta'$ is optimized by a specific algorithm. 

Furthermore, it is necessary to explain the relation between preference learning for LLMs and some related concepts, based on this definition. (1) Alignment: Following \citet{kenton2021alignment}, alignment refers to \textit{the research focuses on tackling the so-called behavior alignment problem: How do we create an agent that behaves in accordance with what a human wants?} Based on this definition, we regard preference learning for LLMs as a category of methods aimed at achieving alignment. The scope of this paper is confined to textual preference alignment which doesn't contain other well-known alignment topics such as \textit{hallucination}, \textit{multi-modal alignment}, and \textit{instruction tuning}.
(2) Reinforcement Learning from Human Feedback (RLHF): Different from RLHF, the scope of this paper not only includes RL-based methods but also encompasses the traditional called SFT-based methods. What's more, we adopt a unified perspective to investigate both reinforcement-learning and supervised-learning-based methods.

\section{The Unified View of Preference Learning for LLM}
Inspired by recent works \cite{shao2024deepseekmath, guo2024direct}, we survey existing works from a unified perspective in the following two folds: 

\textbf{First, the optimization objectives of RL and SFT-based methods can be described within the same framework.} Following \cite{shao2024deepseekmath}, the gradient with respect to the parameter $\theta$ of a training method can be written as:

\begin{equation}
\nabla_{\theta} = \mathbb{E}_{[(q,o) \sim \mathcal{D}]}\left( \frac{1}{|o|} \sum\limits_{t=1}^{|o|} \delta_{\mathcal{A}(r, q, o, t)} \nabla_{\theta}\log \pi_{\theta}(o_t | q, o_{<t})\right),
\label{eq:objective}
\end{equation}

where $\mathcal{D}$ denotes the data source which contains the input question $q$ and the output $o$. $\delta$ denotes the gradient coefficient which directly determines the direction and step size of the preference optimization. $\mathcal{A}$ denotes the algorithm. The gradient coefficient is determined by the specific algorithm, data, and corresponding feedback. Tracing back to one significant source influencing the gradient coefficient is the feedback. Note that the feedback could take on various forms. For example, the correctness of data in RFT \cite{rft} or the preference label in DPO \cite{rafailov2024direct} could impact the gradient coefficient, thereby affecting the final gradient. Consequently, we define the feedback in our paper as the preference given by the environment that can affect the gradient coefficient. Notably, both RL-based and SFT-based methods can be encapsulated within this framework. 

\textbf{Second, the algorithm can be decoupled from online/offline settings.} 
In the context of alignment, online learning refers to the preference oracle $r$ or its approximator $\hat{r}$ can be queried over training, i.e. the feedback of the responses sampled from the current actor model can be given on the fly. 
If the feedback signal cannot be obtained in real-time, then it is considered offline learning.

From a traditional perspective, RL-based methods are more flexible with online/offline settings, while SFT-based methods are typically offline. However, as in the first point where we have unified RL-based and SFT-based approaches, it can be inferred that SFT-based methods can also be applied in an online setting, which has been proved by recent work \cite{guo2024direct}. 
Actually, what determines whether the setting is online or offline is merely whether the preference signal is generated in real-time or pre-stored. 
In section \ref{sec:preference_data}, we elucidate the methods of acquiring data for both online and offline settings. In online settings, data collection typically follows an on-policy strategy, while in offline settings, it generally adheres to an off-policy strategy. Although it is feasible to combine online feedback collection with an off-policy strategy, such instances are relatively rare in the existing works.
Therefore, unlike the categorization in other survey papers \cite{wang2023aligning, shen2023large, wang2024essence}, we do not use online/offline nor RL/SFT as criteria for classifying algorithms. 
Instead, we decouple the algorithm from the online/offline setting. 
As an example, the DPO algorithm does not necessarily have to be offline. 
It depends on the context in which they are actually applied. 
If there is an evaluator available to assess preference relations in real-time generated data, then DPO can also be utilized for online optimization.

Based on the two points discussed above, we ultimately divide preference learning into four key elements: \textbf{Model}, \textbf{Data}, \textbf{Feedback}, and \textbf{Algorithm}, as shown in Figure \ref{fig:formulation}.

The process of preference learning can be described as follows: For a LLM $\pi_\theta$ to be aligned, we first need to prepare the data $\mathcal{D}$ for training. 
If we are in an online setting, we must sample behavioral data from the model and the environment will provide a preference feedback signal to the data in real-time. 
If it is an offline setting, we need to have a preference dataset prepared in advance. 
Whether it conforms to human preference will be reflected in the feedback $\mathcal{R}$. 
For example, in the DPO series of methods, data that do not meet the preferences will be assigned a bad label. 
In RFT, it will be discarded, that is, the gradient coefficient will be zero. 
For RL-based algorithms like PPO, this would correspond to a lower reward score. 
Subsequently, the tuple $(\mathcal{D}_{\{x, y\}}, \mathcal{R}, \pi_\theta)$ is fed into the algorithm $\mathcal{A}$. We categorize algorithms into four types based on the data required for each model update by the algorithm: training-free methods, point-wise methods, pair-wise contrasts, and list-wise contrasts, without the need to be concerned whether they are RL or SFT-based algorithms. 
Finally, we acquire an aligned LLM $\pi_\theta'$. The formal description of this process is provided in Algorithm \ref{algorithm:preference_learning}.

\begin{algorithm}
\caption{Preference Learning} 
\label{algorithm:preference_learning} 
\SetAlgoLined 
\SetKwInOut{Input}{Input} 
\SetKwInOut{Output}{Output}

\Input{$\pi_\theta$ (Initialize LLM to be aligned), $\mathcal{E}$ (Environment with human preference), \\ $\mathcal{Q}$ (Unlabeled queries) or $\mathcal{D}$ (Pre-prepared offline dataset), $\mathcal{A}$ (Algorithm)}
\Output{$\pi_\theta'$ (Aligned LLM)}
\If{reference model is needed}{
    $\pi_{ref} \gets \pi_\theta$\;
}

\While{(Total training steps not reached)}{
    \If{online setting}{
        $\mathcal{B}$ $\gets$ Sample response from $\pi_\theta$\ using $\mathcal{Q}$\text{;} \\
        $\mathcal{R}$ $\gets$ Get the feedback from the environment $\mathcal{E}$ in real time\;
    }
    \ElseIf{offline setting}{
        $\mathcal{B} \gets$ Get a batch of data with the preference feedback from the pre-stored $\mathcal{D}$\;
    }
    $\pi_\theta' \gets$ Feed $(\mathcal{B}_{\{x, y\}}, \mathcal{R}, \pi_\theta, \pi_{ref})$ into $\mathcal{A}$ and update model\;
    $\pi_\theta \gets \pi_\theta'$\;
}

\Return $\pi_\theta$\; \Comment{Return the aligned LLM}

\end{algorithm}

\section{Preference Data}
\label{sec:preference_data}

The preference data does not have a fixed form and we use the simplest notation to represent preference data as $(x, y, r)$. Here $x, y$ are the literal information input and the candidate output. $r$ is a preference label given by certain feedback systems, which could be human, reward models, or other scoring systems.

Current LLM preference learning methods gather the training data from two sources: on-policy or off-policy.
Generally speaking, the on-policy data collection means we collect the data directly from our policy LLM $\pi_{\theta^t}$ at each training step $t$.
The off-policy data collection could be done outside the box, independent from the LLM to conduct preference learning and result in a dataset consisting of data that is not generated by the policy model itself. 
Notably, using preference data sampled from $\pi_{\theta^0}$ to train $\pi_{\theta^t}$ for $t > 0$ is also off-policy.

\subsection{On-policy Data Collection} 
On-policy data collection process is similar to the setting of on-policy reinforcement learning, where the preference data is obtained directly during training: it first samples a batch of experience by the policy LLM and then obtains the reward by interacting with the environment and finally uses it to update the policy LLM. Under such conditions, different methods vary in the preference generator $g(x)$ from the environment.

\paragraph{On-policy Sampling methods} To sample various experiences from the environment, numerous research studies have explored different strategies for decoding. 

Diverse sampling strategies such as Top-K/Nucleus Sampling~\cite{holtzman2020curious} and Beam Search~\cite{graves2012sequence} are employed during the generation process of LLMs. These methods determine the efficiency and effectiveness of the data used for preference learning.

For problems that involve multi-step solutions, there has also been some research~\citep{wang2024mathshepherd, zhang2024accessinggpt4levelmathematical, zhang2024restmctsllmselftrainingprocess, feng2024alphazeroliketreesearchguidelarge, luo2024improvemathematicalreasoninglanguage, qi2024mutualreasoningmakessmaller, zhu2024recoveringmentalrepresentationslarge, xiong2024watchstepllmagent} employing Monte Carlo tree search (MCTS)~\cite{KocsisMCTS} to enhance the diversity and performance of the data sampling. 
MCTS originated from the advancements made in AlphaGo. The fundamental concept of MCTS involves evaluating various strategies through numerous simulations, or rollouts, to determine which strategy yields superior results. This approach resembles a methodical and deliberative thinking process, contrasting with greedy decoding methods that prioritize immediate gains. The core operations of MCTS can be categorized into four distinct phases: \textbf{selection}, \textbf{expansion}, \textbf{simulation}, and \textbf{backpropagation}.
MCTS's efficient search strategy enables models to generate higher-quality data while simultaneously acquiring step-level labels. This refined data can subsequently be utilized to enhance model performance during decoding~\citep{qi2024mutualreasoningmakessmaller, zhu2024recoveringmentalrepresentationslarge,feng2024alphazeroliketreesearchguidelarge}, train reward models~\citep{wang2024mathshepherd, luo2024improvemathematicalreasoninglanguage}, and fine-tune the model~\citep{zhang2024accessinggpt4levelmathematical, zhang2024restmctsllmselftrainingprocess, feng2024alphazeroliketreesearchguidelarge}. 


\subsection{Off-policy Data Collection}
\label{sec:offline_data}
Off-policy data collection entails gathering training data independently of the LLM's learning process. 
This method is generally easier than on-policy data collection, largely due to the availability of open-source preference datasets. 
Alternatively, we can also compile a dataset in advance using the initial model $\pi_{\theta^0}$.
The off-policy data collection strategy ensures a more diverse training dataset that can often yield improvements in the LLM's preference learning process. 
There are two main sources of preference data, the ones from human annotators and those generated by more advanced LLMs. Please note that as relevant research continues to advance, the number of open-source datasets related to preference learning is increasing. Consequently, it is challenging to compile a comprehensive list of all datasets. Therefore, we will only highlight a few representative works.

\paragraph{Data from Human}
Webgpt~\cite{nakano2021webgpt} has 20K comparisons, where each example consists of a question, a pair of model answers, and human-rated preference scores for each answer.

OpenAI’s Human Preferences~\cite{ouyang2022training} originates from a carefully selected portion of Reddit's TL;DR corpus \cite{volske-etal-2017-tl}. Each entry within this dataset consists of a post, paired with two alternative summary options, and is augmented by an evaluation from a human annotator who designates the preferred summary of the two. 

HH-RLHF~\cite{bai2022training} involves 170K chats between humans and AI helpers. In these chats, the AI gives two different replies. A human annotator marks which reply is better and which is not as good. 

SHP~\cite{pmlr-v162-ethayarajh22a} consists of 385K human preferences regarding responses to questions in 18 subject areas, reflecting user preferences for helpfulness. In contrast to HH-RLHF, SHP relies solely on human-written data, allowing for complementary distributions between the two datasets.

\paragraph{Data from LLMs} Obtaining preference from human could be resource-consuming. However, researches~\citep{lee2023rlaif, cui2023ultrafeedback, lyu2024macpo} show that strong LLMs excel at simulating human preferences. Consequently, numerous efforts have been made to utilize LLMs as preference data generators for scaling up.

RLAIF~\cite{lee2023rlaif} curates a comprehensive dataset that amalgamates the Reddit TL;DR corpus~\cite{volske-etal-2017-tl}, OpenAI's Human Preferences~\cite{stiennon2022learning}, and the HH-RLHF~\cite{bai2022training} dataset, the preferences of which are annotated using PALM 2 instead of human. The results of the experiments indicate that scaling up using AI feedback significantly enhances the model's training performance.

Open-Hermes-Preferences~\cite{open_hermes_preferences} is a comprehensive dataset containing roughly 1 million AI-generated preferences. It integrates outputs from this dataset and two additional models and PairRM~\citep{jiang2023llmblender} is employed as the preference model for evaluation and ordering of responses.

ULTRAFEEDBACK~\cite{cui2023ultrafeedback} employs GPT-4 to develop ULTRAFEEDBACK, an expansive, superior-quality, and varied preference dataset designed to overcome the scarcity and constraints of existing preference data.

UltraChat~\cite{ding2023enhancing} is a million-scale multi-turn instructional conversation dataset. Unlike datasets built around specific tasks, UltraChat encompasses a wide array of human-AI interaction scenarios. It leverages advanced techniques like meta-information, in-context expansion, and iterative prompting, alongside two separate ChatGPT Turbo APIs for realistic and informative conversation generation.

\section{Feedbacks}
\label{sec:reward_function}


\begin{figure}[!ht]
    \centering
    \includegraphics[width=1\linewidth]{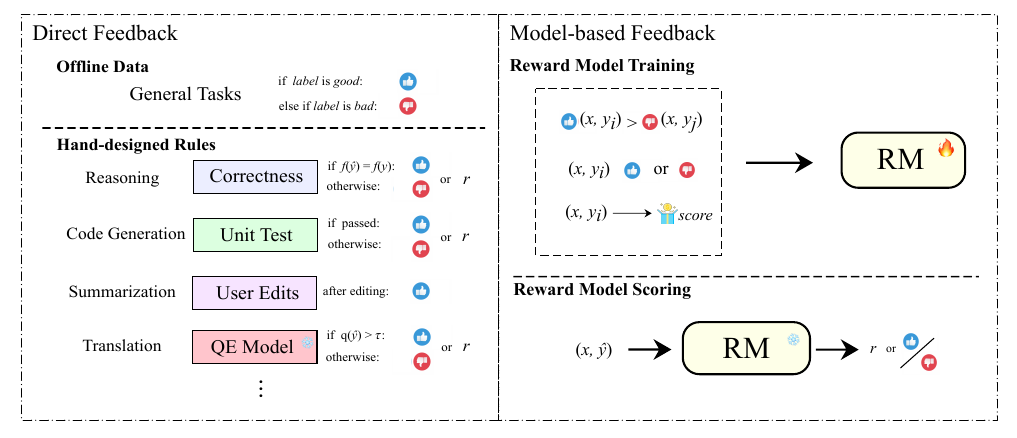}
    \caption{The illustration of the reward received by the model during the preference learning. For a data sample $(x, \hat{y})$, where the $\hat{y}$ is unlabeled candidate output, the reward function is supposed to provide the feedback, which can be the reward score $r$ or the preference label. According to whether we need to train a specific reward model, the reward function can be categorized into \textbf{direct feedback} and \textbf{model-based feedback}.}
    \label{fig:reward_figure}
\end{figure}

In this section, we elaborate on the preference feedback received by the model in preference learning. Following \citet{shao2024deepseekmath}, the feedback in this paper refers broadly to the preference indicators that can influence the gradient of the model during the training process. Herein, it can not only serve exactly as the reward in the methods using reinforcement learning but also be preference labels or other feedback utilized by algorithms that do not explicitly employ reinforcement learning. 
Formally, given a data instance $(x, \{\mathbf{\hat{y}\}})$, where $\{\mathbf{\hat{y}\}} = {\hat{y}_1, \hat{y}_2, ..., \hat{y}_i}$ and $i \geq 1$, the environment aligned with human preference is supposed to give out the reward, which could be preference $y_i > y_j$ or a scalar $r$. As shown in Figure \ref{fig:reward_figure}, we survey various types of feedback in preference learning, categorizing them into two classes: direct feedback, and model-based feedback.

\subsection{Direct Feedback}
Direct feedback refers to the feedback that can be directly obtained without training a specific reward model. 


\paragraph{Labeled Datasets} One of the most direct ways to obtain feedback is through labeled datasets annotated by humans. The labeled preferences within the dataset can be directly utilized for model training in offline methods. 
We cover the recent advancement of the existing datasets on preference learning in Section \ref{sec:offline_data}. 

\paragraph{Hand-designed Rules} The other way to obtain direct reward is to use hand-designed rules as a reward. Due to the specificity of the rules, it is difficult to establish a unified criterion that encompasses all methods. Different tasks may adhere to different sets of rules. 

For the task of mathematical reasoning, \citet{rft} utilize the correctness of the reasoning paths as a metric to control the training data. 
Following \citet{shao2024deepseekmath}, which provides another perspective of these series of methods, the reward can be calculated by $r=\mathbb{I}(c)$, where the reward equals to 1 if the COT reasoning path is correct and 0 otherwise.
\citet{xin2024deepseekproveradvancingtheoremproving, xin2024deepseekproverv15harnessingproofassistant} obtaining the feedback of the math theorem prover using the automated proofing tool (LEAN~\citep{LEAN_4}).  
For machine translation, \citet{xu2024contrastive} utilize the results of the reference-free QE model to obtain the preference of different translation candidates and further optimize the model using CPO, an improvement of the DPO algorithm. 
For code generation, \citet{shen2023pangucoder2} rank the model output according to unit tests and heuristic preferences. For each data, they assign different scores from low to high based on the different situations of the test results. The preference obtained from the current ranking directly affects the final training loss of the model. \citet{liu2023rltf} and \citet{dou2024stepcoder} convert the results of unit tests under different scenarios into scalars using hand-designed rules and further optimize the model using RL algorithms.
For summarization, \citet{gao2024aligning} explores interactive learning for the model by using textual human edits on the agent's outputs, which proved to be simple and effective.

\subsection{Model-based Feedback} 
In this section, we conduct a survey of model-based feedback, encompassing reward signals from reward models, pairwise scoring models, and LLM-as-a-Judge feedback.

\subsubsection{Reward Model}
\label{subsec:rm2}
Training a reward model is predicated on constructing a classifier that can anticipate human preference probability, $p$, between two outputs. 

\paragraph{Bradly-Terry based Reward Model} One line of research models the preference of humans using the Bradley-Terry model\cite{bradley1952rank}. This involves training the model to estimate $p$, derived from the comparison between two potential responses, by maximizing the likelihood of the preferred output. Parameters of this model are optimized through a loss function that emphasizes the difference in preference between a chosen and a rejected output:

\begin{equation}
     p^*(y_{1} \succ y_2 \ | \ x) = \frac{\text{exp}(r^*(x, y_1))}{\text{exp}(r^*(x, y_1)) + \text{exp}(r^*(x, y_2))}.
\end{equation}

The model is typically optimized by a negative log-likelihood loss: 
\begin{equation}
    \mathcal{L}_{r} = -\log \sigma \left(r^*\left(y_c,x\right)-r^*\left(y_r,x\right)\right)
\end{equation}
where $y_r$ represents the rejected output and $y_c$ represents the chosen output. At inference time, the reward model returns a scalar $p^*(y_{1} \succ y_2 \ | \ x)$ which represents the probability that the output would be the preferred response.

\paragraph{Binary Classifier based Reward Model} For tasks where the quality of a case can be determined by its outcomes, directly labeling samples to train a binary classifier as a reward model is a straightforward and stable approach. For instance, in mathematical reasoning, a sample can be labeled based on whether the response yields the correct final answer. Similarly, in code generation tasks, labeling can be done by checking if the generated code passes specified tests. Unlike tasks such as text summarization or dialogue generation, which require pairwise comparisons of examples, these direct assessment methods simplify the preference labeling process.

Unlike the traditional Bradley-Terry Reward Model, once the labels for the data are obtained, the reward model can be trained using point-wise binary classification loss without needing to construct pairwise data. The BCE training loss is as follows:
\begin{equation}
    L_{r}=-[rlog(\hat{r})+(1-r)log(1-\hat{r})]
\end{equation}

where $r$ is the preference label and $\hat{r}$ is the predicted reward.

\paragraph{RM Training Optimization}
In pursuit of a better reward model, numerous studies optimize the existing reward models from various perspectives. 

One line of research seeks to obtain better preference data. \citet{lee2023rlaif} capitalizes on the capabilities of off-the-shelf LLMs to generate preference labels, potentially decreasing the need for expensive and time-consuming human annotation. The study showcases that RLAIF can reach or even surpass the performance levels of RLHF across multiple tasks. \citet{jinnai2024regularized} explore using  Kullback–Leibler divergence and Wasserstein Distance to regularize the Best-of-N sampling, which proved to be effective in mitigating the reward hacking problem during reward modeling. \citet{pace2024westofn} utilizes West-of-N to generate better synthetic preference data, extending Best-of-N sampling strategies from language model training to the reward model training. 

Another line of research focuses on model ensembling to improve the overoptimization and uncertainty estimation of the reward model. \citet{coste2024reward} use reward model ensemble to mitigate the reward model overoptimization. \citet{quan2024dmoerm} propose a novel and effective reward modeling method based on MoE, which decomposes the inputs into different capability points under different tasks. \citet{zhai2023uncertaintypenalized} consider LoRA-based ensemble, while their work focuses on an uncertainty penalized objective in RL-finetuning. \citet{ramé2024warm} consider a different approach of averaging the weights of multiple reward models instead of ensembling their predictions. \citet{zhang2024improving} explore multiple ensembling methods for developing efficient ensemble approaches. 

Exploring another dimension, research on fine-grained rewards is gaining momentum. \citet{NEURIPS2023_b8c90b65} introduce Fine-Grained RLHF, a framework that enables training and learning from reward functions that provide rewards in multiple aspects after every segment. \citet{yang2024regularizing} focuses on optimizing the training process of reward models through text-generation regularization. In contrast to outcome supervision, which provides feedback for a final result, \citet{uesato2022solving},  \citet{lightman2023lets} and \citet{yu2024ovm} explore process supervision, which provides reward for each intermediate reasoning step. However, the training data for PRM is constrained by the high effort required for annotation, and how to efficiently construct step-level training data remains a challenge. \citet{wang2024mathshepherd} construct process supervision data in an unsupervised manner, which proved to be effective for mathematical reasoning. 

Additionally, optimizing the training process of reward models is a focus area, \citet{zhou2024prior,dong2024rlhf} proposes using prior constraints to mitigate the uncontrolled scaling of reward scores during training the reward model. \citet{gao2024llmcriticshelpcatch} proposes a two-stage training paradigm, utilizing natural language feedback to stimulate the evaluation ability of the mathematical reward model.

\subsubsection{Pair-wise Scoring Model}
In addition to specially trained reward models, lightweight pairwise scoring models are widely used to provide preference signals for models~\cite{jiang2023llmblender}. Generally speaking, pairwise scoring models employ a specialized pairwise comparison method to distinguish subtle differences between candidate outputs. Since it is easier and more consistent to discriminate among multiple candidates rather than scoring individual candidates each time, pairwise scoring models are usually smaller and achieve better results. For instance, the PairRanker~\cite{jiang2023llmblender}, with only 0.4B parameters, exhibits the highest correlation with ChatGPT-based ranking and is widely used in works such as SPPO~\cite{sppo} and SimPO~\cite{meng2024simpo}. However, pairwise methods cannot provide a global score, and the number of candidates they can process at one time is limited. Consequently, obtaining a global ranking among multiple candidates or a general reward signal often incurs a higher cost.

\subsubsection{LLM-as-a-Judge}
A more direct and easily adjustable approach is to use LLM scoring to provide rewards for preference learning or evaluation, termed LLM-as-a-Judge. For larger models, such as GPT-4, we can specify scoring rules directly in the prompts, allowing the model to score generated responses. Extending this method further, we can implement LLM self-rewarding. For example, recent self-rewarding mechanisms~\cite{yuan2024self} have shown that LLMs can improve by evaluating their own responses instead of relying on human labelers. However, model judgment may introduce errors or biases. To address this issue, \citet{wu2024meta} introduce a novel Meta-Rewarding step, where the model assesses its own judgments and uses that feedback to refine its judgment skills. This unsupervised approach makes the scores given by the LLM more accurate. To address extreme multi-objective optimization, \citet{xu2024perfect} employs Mixture of Judges with cost-efficient constrained policy optimization with stratification, which can identify the perfect blend in RLHF in a principled manner. \citet{ye2024beyond} uses the LLMs self-generated contrastive judgment pairs to train the generative judge with DPO and find the performance is comparable to the scalar reward model. For tasks involving complex reasoning steps, LLM-as-a-Judge often underperforms the trained scoring verifiers. To mitigate this issue, \citet{mcaleese2024llmcriticshelpcatch} train a critic model prompted to accept a (question, answer) pair as input and output a plain text “critique” that points out potential problems in the answer for code generation. \citet{zhang2024generativeverifiersrewardmodeling} train a generative verifier to leverage the text-token prediction capabilities of LLMs for mathematical reasoning.

\section{Algorithms}
\label{sec:algorithms}
Preference learning algorithms optimize LLMs to align with human preferences based on data and feedback. Based on the number of samples needed to compute the gradient coefficient in formula~\ref{eq:objective}, we can categorize these algorithms into three types: point-wise methods, pair-wise contrasts, and list-wise contrasts. Point-wise methods rely on the quality of a single sample to determine the gradient coefficient, pair-wise contrasts require comparisons between pairs of samples, and list-wise contrasts involve evaluating entire lists of samples to compute the gradient coefficient. In addition to this, there are also algorithms that optimize the models without the need for training, we categorize them into training-free alignment.

In summary, we categorize the preference algorithms into four groups: point-wise methods, pair-wise contrasts, list-wise contrasts, and training-free alignment. For some representative algorithms belonging to pair-wise/list-wise contrasts, we also provide their detailed loss designs in Table~\ref{loss}.

\subsection{Point-wise Method}

Point-wise methods optimize the model based on a single data point $(x, y)$. 
Due to the fact that these methods do not require paired preference data during optimization, the cost of labeling preference data is significantly reduced.
The point-wise methods are easy to implement and have demonstrated effectiveness in a series of situations \citep{rft, star, raft,schulman2017proximal,shao2024deepseekmath,li2023remax, ethayarajh2024kto}.


The simplest point-wise optimization method is the rejection sampling fine-tuning. 
This approach first selects high-quality data points using a reward function or rules and then fine-tunes LLMs on these selected data. The object of rejection sampling fine-tuning is shown in \cref{eq:rs_obj}, where $y^{+}$ is the response with high reward.
\begin{equation}
\label{eq:rs_obj}
L_{RS}=-\sum_t \log \pi_{\theta}\left(y^{+}_{t} \mid x, y^{+}_{<t}\right)
\end{equation}
There are several works that demonstrate the effect of rejection sampling fine-tuning. 
RAFT \citep{raft} employs a reward model to rank generated samples, filtering out those that best align with human preferences and values. Using these curated samples, the model undergoes fine-tuning to enhance its friendliness and accessibility to humans.
Star \citep{star} iteratively utilizes a limited selection of rationale examples alongside a substantial dataset lacking rationales, enhancing the capacity for increasingly complex reasoning without relying on a reward model.
Yuan et al. \citep{rft} employ rejection sampling fine-tuning to enhance the mathematical reasoning capabilities of LLMs, as they discover that the chosen samples encompass a greater variety of distinct reasoning paths, which proves advantageous for tackling math problems.
Although simple and straightforward, rejection sampling fine-tuning fails to leverage the data with low reward, which prevents it from learning from non-preferred data and optimizing the model further.

Proximal Policy Optimization (PPO) \citep{schulman2017proximal} from OpenAI is one of the most representative and successful point-wise optimization algorithms.
Notably, the most successful applications like ChatGPT and GPT-4 are produced by the PPO method. 
During the PPO optimization phase, we update the LM to maximize the return from the learned reward function $r$ using the following principle:
\begin{equation}
  \label{eq:ppo_obj}
    \max_\theta J_r(\theta) = \max_\theta \sum_{x} \mathbb{E}_{a \sim \pi_\theta(\cdot|x)}\left[r(x, y) - \beta \log \frac{\pi_\theta(y|x)}{\pi_{ref}(y|x)}\right],
\end{equation}
where $\pi_{ref}$ is the supervised fine-tuned model and $\pi_\theta$ is initialized as $\pi_{ref}$. $\beta$ is the KL divergence coefficient, which controls the deviation from the original model.
Despite the impressive performance demonstrated by OpenAI, the PPO algorithm requires extensive computational resources and suffers from sample inefficiency. 
Another drawback of the PPO algorithm is its training instability, making it challenging to determine the appropriate hyperparameters for the PPO method.

To address the drawbacks of PPO outlined earlier, several alternative point-wise methods have been proposed. One such method is ReMax \citep{li2023remax}, which draws on concepts from the REINFORCE algorithm \citep{williams1992simple}. 
Notably, ReMax modify the calculation of gradient coefficient by
incorporating a subtractive baseline value, and the baseline value can be defined as the reward of a greedy sampling response.
The authors suggest that this approach reduces computational requirements by eliminating the need for a critic model, which is essential for PPO, while also promoting more stable training.
From our perspective, ReMax should be "pairwise" as it introduces an additional subtractive baseline for computing the gradient coefficient. We introduce ReMax in this section to help readers understand the process without being too abrupt.

Another point-wise method is KTO from Ethayarajh et al. \citep{ethayarajh2024kto}.
This approach requires very few predetermined hyperparameters, ensuring stable training while also being resource-efficient. 
Due to the adoption of Kahneman \& Tversky’s prospect theory, KTO doesn't need the pair-wise preference dataset. 
It only requires a label denoting whether the response is preferred or not. KTO directly maximizes the utility of generations instead of maximizing the likelihood of preferences.
Leveraging these point-wise preference data, KTO can achieve prior or comparable performance compared with DPO. 
At the same time, the KTO algorithm also demonstrates good performance in cases of extreme data imbalances \citep{ethayarajh2024kto}, which makes it a good choice in certain specific circumstances.
\begin{table*}[t]
\centering
\resizebox{1\textwidth}{!}{
\begin{tabular}{lll}
    \toprule
    \multicolumn{1}{l}{\multirow{1}{*}{\textbf{Algorithms}}} & \multicolumn{1}{c}{\multirow{1}{*}{\textbf{Formulations of Loss Function}}} & \multicolumn{1}{c}{\multirow{1}{*}{\textbf{Notes}}}\\
    \midrule
    \multicolumn{1}{l}{DPO~\citep{rafailov2024direct}} & \multicolumn{1}{c}{$-\log\sigma \left(\beta\log\frac{\pi(y^+\mid x)}{\pi_\text{ref}(y^+\mid x)}- \beta\log\frac{\pi(y^-\mid x)}{\pi_\text{ref}(y^-\mid x)}\right)$} & \multicolumn{1}{c}{\begin{tabular}{c} $\beta$ is a hyperparameter, \\typically retained by \\other DPO-like methods.\end{tabular}}\\
    \midrule
    \multicolumn{1}{l}{IPO~\citep{azar2023general}} & \multicolumn{1}{c}{$\left(\log\frac{\pi(y^+|x)\pi_\text{ref}(y^-|x)}{\pi(y^-|x)\pi_\text{ref}(y^+|x)}-\frac{1}{2\tau}\right)^2$} & \multicolumn{1}{c}{\begin{tabular}{c} $\tau$ is a hyperparameter \\determining the upper bound \\of the scoring margin.\end{tabular}}\\
    \midrule
    \multicolumn{1}{l}{$f$-DPO~\citep{wang2023beyond}} & \multicolumn{1}{c}{$-\log\sigma \left[\beta f'\left(\frac{\pi(y^+\mid x)}{\pi_\text{ref}(y^+\mid x)}\right)- \beta f'\left(\frac{\pi(y^-\mid x)}{\pi_\text{ref}(y^-\mid x)}\right)\right]$} & \multicolumn{1}{c}{\begin{tabular}{c} $f'$ is the derivative of\\the chosen $f$-divergence.\end{tabular}}\\
    \midrule
    \multicolumn{1}{l}{EXO~\citep{ji2024towards}} & \multicolumn{1}{c}{$-\sum_{y\in(y_w,y_l)} \frac{\left(\frac{\pi(y|x)}{\pi_\text{ref}(y|x)}\right)^{\beta_\pi}}{\sum_{y'\in(y_w,y_l)}\left(\frac{\pi(y'|x)}{\pi_\text{ref}(y'|x)}\right)^{\beta_\pi}}\log\frac{\left(\frac{\pi(y|x)}{\pi_\text{ref}(y|x)}\right)^{\beta_\pi}/|\mathbf{1}_{\{y = y_w\}}-\epsilon|}{\sum_{y'\in(y_w,y_l)}\left(\frac{\pi(y'|x)}{\pi_\text{ref}(y'|x)}\right)^{\beta_\pi}}$} & \multicolumn{1}{c}{\begin{tabular}{c} $\beta_\pi$ and $\epsilon$ are hyperparameters.\\$\epsilon$ is used to soften the binary \\human-crafted rewards.\end{tabular}}\\
    \midrule
    \multicolumn{1}{l}{DPO-positive~\citep{pal2024smaug}} & \multicolumn{1}{c}{$-\log\sigma \left(\beta\log\frac{\pi(y^+\mid x)}{\pi_\text{ref}(y^+\mid x)}- \beta\log\frac{\pi(y^-\mid x)}{\pi_\text{ref}(y^-\mid x)}\right)-\lambda\max\left(0, \frac{\pi_\text{ref}(y^+|x)}{\pi(y^+|x)}\right)$} & \multicolumn{1}{c}{$\lambda$ is a hyperparameter.}\\
    \midrule
    \multicolumn{1}{l}{ORPO~\citep{hong2024orpo}} & \multicolumn{1}{c}{$\mathcal{L}_\text{sft} -\lambda\log\sigma\left[\log\frac{\pi(y^+|x)(1-\pi(y^-|x))}{\pi(y^-|x)(1-\pi(y^+|x))}\right] $} & \multicolumn{1}{c}{$\lambda$ is a hyperparameter.}\\
    \midrule
    \multicolumn{1}{l}{SimPO~\citep{meng2024simpo}} & \multicolumn{1}{c}{$-\log\sigma\left[\frac{\beta}{|y^+|}\log\pi(y^+|x)-\frac{\beta}{|y^-|}\log\pi(y^-|x)-\gamma\right]$} & \multicolumn{1}{c}{\begin{tabular}{c} $\gamma$ is a hyperparameter to\\control the margin between\\the scores of $y^+$ and $y^-$.\end{tabular}}\\
    \midrule
    \multicolumn{1}{l}{RRHF~\citep{yuan2024rrhf}} & \multicolumn{1}{c}{$\mathcal{L}_\text{sft} + \sum_{i>j}\max\left[0, \pi(y^i|x)-\pi(y^j|x)\right]$} & \multicolumn{1}{c}{-}\\
    \midrule
    \multicolumn{1}{l}{PRO~\citep{song2023preference}} & \multicolumn{1}{c}{$\beta\mathcal{L}_\text{sft} - \sum_{k=1}^{n-1} \log \frac{\exp\left[\frac{\pi(y^k|x)}{1/(r^*(x, y^k) - r^*(x, y^n))} \right]}{\frac{\pi(y^k|x)}{1/(r^*(x, y^k) - r^*(x, y^n))} + \sum_{i=k+1}^{n}\exp\left[ \frac{\pi(y^i|x)}{1/(r^*(x, y^k) - r^*(x, y^i))} \right]}$} & \multicolumn{1}{c}{\begin{tabular}{c}$\beta$ is a hyperparameter to\\balance $\mathcal{L}_\text{sft}$ and the rest\\of list-wise contrasts.\\$r^*$ is an external reward model.\end{tabular}}\\
\bottomrule
\end{tabular}
}
\caption{\label{loss}
    Demonstrations of loss function design for different algorithms. Due to the limitation of page width, there are just several algorithms selected here for they can be placed in one line. Therefore, we strongly recommend the direct reference of their papers for more works mentioned in this survey. 
}
\end{table*}

\subsection{Pair-wise Contrast}
\citet{liu2024chain} point out that point-wise methods either solely rely on the positive candidates, learning to conduct mindless intimation instead of truly understanding human preference against those negative candidates, or they face significant optimization challenges, complicating their practical application.
As a result, they propose Chain-of-Hindsight~(CoH) where a pair of positive and negative candidates $y^+$ and $y^-$ are both placed in the context during fine-tuning, accompanied by corresponding prompts. This helps the tuned LLM learn from semantic contrasts. For inference, the LLM is triggered by the \textit{positive} prompt~(e.g. \textit{A helpful response is:}) to generate preferred responses.

However, such prompt methods are insufficient to force LLM to sense distinctions about human preference. Instead, more researchers choose to manipulate its inner states~(e.g. probability of generating candidates) to explicitly construct pair-wise contrastive learning. 
For instance, \citet{zhao2023slic} utilize the formulation of SLiC~\citep{zhao2022calibrating} to apply pairwise contrasts between $y^+$ and $y^-$. One of the highlights is their expansion of the training dataset to include additional candidate $y$ sampled from the initial SFT checkpoint. Another representative method is Direct Preference Optimization~(DPO) designed by \citet{rafailov2024direct}. They change the objective of RLHF as an equation between the given reward model $r$, reference model $\pi_\text{ref}$, and corresponding optimal policy $\pi_r$,
\begin{equation}
\label{dpo1}
    r(x,y) = \beta \frac{\pi_r(y\mid x)}{\pi_\text{ref}(y\mid x)} + \beta \log Z(x)
\end{equation}
where $x$ and $y$ are the context and its candidate response, respectively. Combining Equation~\ref{dpo1} with the Bradley-Terry~\cite{bradley1952rank} loss used in reward models, DPO formulates a new objective for direct optimization of policy $\pi$ while containing an implicit reward learning process,
\begin{equation}
\begin{aligned}
    \mathcal{L}_\text{DPO}(\pi;\pi_\text{ref})
    =&-\log\sigma (\beta\log\frac{\pi(y^+\mid x)}{\pi_\text{ref}(y^+\mid x)}-\beta\log\frac{\pi(y^-\mid x)}{\pi_\text{ref}(y^-\mid x)})
\end{aligned}
\end{equation}
Despite its effectiveness~\citep{miao2024aligning}, \citet{azar2023general} demonstrates that DPO can easily overfit the pair-wise annotations from the provided preference datasets. 
It arises from that DPO transforms the score of $y^+$ to an unbounded range using a non-linear mapping $\psi(q)=\log\frac{q}{1-q}$. This leads to an extreme optimization of the score difference while reducing the impact of the regularization term in RLHF simultaneously.
The authors accordingly propose Identity-PO~(IPO) that replaces the unbounded mapping with the identity mapping. In essence, IPO constrains the upper bound of the above score difference to alleviate overfitting.

After the appearance of IPO, more researchers attempt to modify $\mathcal{L}_\text{DPO}$ for better performance. 
\citet{wang2023beyond} point out the constraint of reverse KL divergence in RLHF for the diversity of generated content, which can be mitigated by other $f$-divergences. They consequently abstract a general DPO-like loss function, providing a plug-and-play form for different $f$-divergences. 
\citet{ji2024towards} claim that, unlike RLHF, DPO essentially optimizes a forward-KL divergence $\text{KL}(\pi_{\text{ref}}||\pi)$. As a substitute, they directly build the objective equivalent to the reverse-KL divergence by a simple estimation of the partition function. 
\citet{chen2024mallows} and \citet{ramesh2024group} both focus more on the input context. 
\citet{chen2024mallows} rely on the condition of Mallow~\citep{mallows1957non} formulation to model the effect of input context on the finally acquired reward, which replaces the original reward modeling in DPO, while \citet{ramesh2024group} utilize the context to place fine-grained controlling information. 
Moreover, some researchers find that DPO tends to reduce the log-likelihood of $y^+$, which could encourage LLM to generate sub-optimal responses \citep{feng2024towards,yan20243d}. \citet{pal2024smaug} then proposed the DPO-positive method that appends a penalty term to mitigate this phenomenon. \citet{yu2024direct} also takes the way of appending penalty term, which, however, sources from prompting the LLM itself. 

Another direction is the modification of the training pipeline. Typically, DPO involves two progressive stages: the first is SFT, and the next is contrastive alignment. Hence, one way of the modification observed is to reduce the cost of fine-tuning. \citet{hong2024orpo} append a term of Odds Ratio to the initial SFT loss for enhanced supervision. This design substantially follows the spirit of pair-wise contrasts, while combining the above two stages into one to shorten the pipeline. 
Besides, a concise DPO-like algorithm has recently been proposed, named SimPO\citep{meng2024simpo}. 
It inherits the framework of DPO, but eliminates the reference model $\pi_\text{ref}$ in the second stage to directly optimize the log-likelihood of $y^+$ exceeding other candidates, which, on the other hand, aligns with the nature of maximized log-likelihood of sequences in LLM inference.
Other attempts can be transferring the offline DPO into online settings. For example, \citet{kim2024sdpo} and \citet{gorbatovski2024learn} share a motivation of dynamically updating $\pi_\text{ref}$ in DPO, including full replacement and soft merging. \citet{kim2024sdpo} provide convincing evidence of this manner: through transforming its loss, DPO can be viewed as optimizing $\pi$ to keep $\frac{\pi(y^+|x)}{\pi(y^-|x)}$ away from $\frac{\pi_\text{ref}(y^+|x)}{\pi_\text{ref}(y^-|x)}$, and updating $\pi_\text{ref}$ iteratively force $\pi$ to converge to the optimal policy. \citet{morimura2024filtered} share the idea of online data collection and selection with \citet{raft}, but leverage it on DPO. This process aims to alleviate the effect of original low-quality data. \citet{liu2023statistical} and \citet{zhang2024ts} complete similar targets with more complex frameworks.

Some works also promote the application of preference alignment.
For instance, \citet{hejna2023contrastive} derives an offline policy LLM learning method based on the regret-based model of human preference, named Contrastive Preference Learning, which can model the preference in more complex scenarios, like robotics. With careful re-definitions of preference in specific domains, \citet{she2024mapo} and \citet{lyu2024knowtuning} successfully transfer DPO to multilingual reasoning and knowledge-aware QA, while \citet{zhou2023beyond}, \citet{guo2024controllable}, \citet{badrinath2024hybrid}, \citet{yang2024rewards}, \citet{yang2024metaaligner} and \citet{wang2024conditioned} focus on the adaptation of multi-objective preference alignment.

\subsection{List-wise Contrast}
Extending pair-wise contrasts to list-wise ones is also a natural inspiration, whose effectiveness has been demonstrated by \citet{song-etal-2024-scaling}. 
RRHF~\citep{yuan2024rrhf} pioneers the early adoption of expanding two candidates ${y^+, y^-}$ to a longer list ${y_i}$ using external LLM. 
Nevertheless, this method pairs each candidate $\{y^i\}$ with its inferior ones $y^{>i}$ to create multiple pairs, hence maintaining the utilization of pair-wise contrasts.
\citet{song2023preference} further propose Preference Ranking Optimization~(PRO) to recursively applies multiple list-wise contrasts between $y^i$ and $y^{>i}$.
\citet{hong2023cyclealign} then leverage list-wise contrasts on the distillation of superior black-box LLMs, acquiring improved performance.

Vanilla list-wise contrasts may lead to performance degradation due to biased estimation of candidates, requiring more fine-grained design.
\citet{wang2023making} and \citet{mao2024don} implement multiple calibration strategies on computed scores for different ranking objectives to alleviate over-fitting. 
Differently, \citet{liu2024lipo} and \citet{zhu2024lire} opt to design re-weighting mechanisms on each term in loss functions with external scoring information, which are beneficial to precise scoring ability of $\pi$.
Some alternative list-wise methods have introduced enhancements to the currently most successful PPO algorithm.
For instance, GRPO \citep{shao2024deepseekmath} samples a list of responses for a given query and employs a reward model to evaluate each output. The reward for each response is normalized by subtracting the average reward of the response list and dividing it by the list's standard deviation.
For each output, GRPO sets the advantage to the normalized reward, eliminating the use of the critic model that PPO relies on, thereby reducing the consumption of storage resources during training.

\subsection{Training-Free Alignment}
Training-free alignment refers to approaches that do not fine-tune the language model itself, by which parameters of the language model remain unchanged after alignment. Instead, training-free alignment enhances the model output to better align with preferences by optimizing input prompts (\S \ref{subsection: input optimization}) or optimizing at the output stage (\S \ref{subsection: output optimization}). Optimization on the input side includes incorporating in-context learning examples \citep{song2024icdpo,lin2023unlocking}, retrieval-augmented content \citep{xu2023align} into the prompts, and employing a prompt rewriter \citep{cheng2023black} to refine the prompts before fed into the model.
On the output side, optimization involves redistributing the model's output probability distribution over vocabulary \citep{yang2021fudge,deng2023reward,huang2024deal}, backtracking and regenerating when harmful content is encountered during decoding \citep{li2023rain}, and adding a module after the model to rewrite the originally generated content \citep{ji2024aligner}.

\subsubsection{Input Optimization}
\label{subsection: input optimization}
\citet{lin2023unlocking} analyzes the token distribution shifts between the content generated by base models and aligned models, finding that most shifts occurred with stylistic tokens. Building on this insight, they employ system prompts and restyle the responses of in-context learning examples to align the model.
\citet{xu2023align} aligns the generated contents by retrieving norms most relevant to the given prompt.
BPO \citep{cheng2023black} posits that an effective prompt can lead to better responses. Based on this concept, BPO trains a sequence-to-sequence model to act as a prompt rewriter by using low-quality and high-quality prompt pairs generated by ChatGPT and uses it to optimize the input during inference. 

\subsubsection{Output Optimization}
\label{subsection: output optimization}
\paragraph{Paraphrasing}
Aligner \citep{ji2024aligner} trains an additional alignment module. During inference, the instruction is first fed into the original model to generate an unaligned response; this unaligned response, along with the instruction, is then fed into the alignment module to produce an aligned response.
\paragraph{Logits Manipulation}
FUDGE \citep{yang2021fudge}, \citet{mudgal2023controlled,deng2023reward,liu2024decoding} achieve alignment by modifying the model's output probability distribution during the decoding phase, and they consequently increase the likelihood of generating aligned responses and decrease the probability of harmful responses.
\paragraph{Searching}
RAIN \citep{li2023rain} achieves alignment by rewindable decoding, where harmful tokens are discarded while helpful tokens are preserved, and the quality of the token is evaluated by the LLM itself. 
\citet{huang2024deal} replace the self-evaluation with customized reward models to allow more fine-grained requirements of personal preference in aligned decoding.
ICDPO \citep{song2024icdpo} designs a two-stage Best-of-N-like process to optimize output, where the final response is selected among multiple candidates sampled from a local LLM upon In-context Learning~(ICL), according to their different degrees of human preference. Unlike the traditional Best-of-N that relies on external verifiers (e.g. reward models) for response selection, ICDPO proposes a skillful formulation that aggregates the states before and after ICL as a joint estimation, which is completed solely by the local LLM itself. It can achieve comparable performance as fine-tuning but requires just a few high-quality demonstrations, reducing the implementation cost.

\section{Evaluation}
\label{section:evaluation}
For evaluation of preference learning, the ideal approach is human assessment, such as the Chatbot Arena~\cite{chiang2024chatbotarenaopenplatform}, a benchmarking platform for large language models (LLMs) that facilitates anonymous, randomized matchups through crowd-sourcing. However, due to resource constraints and the potential biases associated with human evaluations, the prevailing method remains automated assessment, which is divided into two parts:
rule-based evaluation as Sec.~\ref{subsection:rule_evaluation} and
LLM-based evaluation as Sec.~\ref{subsection:llm_evaluation}.


\subsection{Rule-based Evaluation}
\label{subsection:rule_evaluation}



Rule-based evaluation is generally conducted in a scheme where the dataset has ground-truth output for each input. In this way, the evaluation can be done by using the widely used automatic metrics including Accuracy, F1, Exact-Match~\cite{rajpurkar2016squad} and ROUGE~\cite{lin2004rouge}.

Current evaluation of general-purpose LLM mainly focuses on evaluating some core tasks before they can be generalized to satisfy various practical needs.
LLMs~\cite{achiam2023gpt,touvron2023llama,touvron2023llama2,vicuna2023} are evaluated on a multi-aspect evaluation set to cove key capabilities.

\paragraph{Factual Knowledge} Factual Knowledge is essential for language models to serve users’ information needs, including Massive Multitask Language Understanding dataset (MMLU)~\cite{hendrycks2020measuring}, C-Eval~\cite{huang2024c} and Massive Multitask Language Understanding in Chinese (CMMLU)~\cite{li2023cmmlu}.

\paragraph{Math} Mathematical reasoning includes the test split of Grade School Math dataset (GSM8K)~\cite{cobbe2021training}, MATH~\cite{hendrycks2020measuring} and Chinese Elementary School Math Word Problems dataset (CMATH)~\cite{wei2023cmath}.

\paragraph{Reasoning} Reasoning is a fundamental ability for LLMs, especially to solve complex problems, including Big-Bench-Hard (BBH)~\cite{suzgun2022challenging}.

\paragraph{Closed-Book Question Answering} Closed-Book question answering includes TriviaQA~\cite{joshi2017triviaqa}, NaturalQuestions~\cite{kwiatkowski2019natural}, CSQA~\cite{saha2018complex} and StrategyQA~\cite{geva2021did}.

\paragraph{Coding} Coding is a special application that people tend to utilize LLMs and might be significant for integrating LLMs with external tools. LLMs would be better at tool usage and function calling with better coding skills. Coding benchmarks includes MBPP~\cite{austin2021program} and HumanEval~\cite{chen2021evaluating}.
Nowadays Coding evaluation benchmarks tend to evaluate LLMs at repo-level code generation including SWE-Bench~\cite{jimenez2023swe} and ML-Bench~\cite{liu2023ml}.

However, the rule-based evaluation strategy with standard metrics suffers from significant drawbacks. The standard metrics only present whether the models' outputs are close to the ground truth outputs, but the currently popular tasks are generally open-ended (i.e., summarization). Calculating standard metrics between models' outputs with the ground truth labels is a misleading evaluation.

\subsection{LLM-based Evaluation}
\label{subsection:llm_evaluation}

With the recent advances of LLMs, LLM-based assistants have started to exhibit artificial general intelligence across diverse tasks (i.e., writing, chatting, and coding). 
Rule-based evaluation of the aforementioned tasks is challenging.
Some recent works~\cite{gudibande2023false, chiang2024chatbot} have found inconsistencies between the performance of language models in human evaluations and NLP benchmark, which may be due to existing evaluation (rule-based evaluation) that only measures LLMs’ core capability on a confined set of tasks (e.g., multi-choice knowledge or retrieval questions) without considering the alignment with human preference in open-ended tasks.
There is an emergent need for a robust and scalable automated method to evaluate LLM alignment with human preferences.
Consequently, Using Large Language Models (LLMs) as proxies for human evaluation to assess the quality of other LLMs has emerged as a cost-effective and promising approach.


\subsubsection{LLM-based Evaluation Methods} LLM-based evaluation methods can be mainly categorized into the following three types:

\paragraph{Pairwise Comparison} LLM evaluators are provided with instructions and the corresponding outputs from two models, then asked to choose the preferred one or declare a tie.~\cite{wang2023large, wang2023automated, li2023generative, wang2023pandalm, kim2023prometheus} Pairwise Comparison is generally the most prevalent method and boasts the highest consistency between human evaluation and LLM-based evaluation, but this method itself is not scalable because the computational costs (.e., the number of possible pairs) will grow significantly as the number of evaluated models increases.
AlpacaEval~\cite{alpaca_eval} is a fast, cost-effective, and reliable LLM-based automatic evaluation system. It uses the AlpacaFarm~\cite{dubois2023alpacafarm} evaluation set, which is designed to assess models' ability to follow general user instructions. Model responses are compared to reference responses using GPT-4-based auto-annotators, producing the win rates presented above. AlpacaEval~\cite{alpaca_eval} exhibits a high level of agreement with human annotations, and its leaderboard rankings strongly correlate with those generated by human evaluators.
AlpacaEval introduces a metric based on the win rate of two LLM-generated responses, as judged by human evaluators. AlpacaEval 2.0~\cite{dubois2024length} further refines this by introducing a length-controlled win rate, which debiases the win rates by accounting for output length differences.
    
\paragraph{Single Answer Grading} Alternatively, an LLM evaluator can be asked to assign an evaluation score to a single instruction and the corresponding answer.~\cite{liu2303g, jain2023multi, kocmi2023large, li2023prd, wang2023automated, kim2023prometheus} Single Answer Grading is efficient for ranking multiple models but fails to discern subtle differences between specific pairs and may show significant score fluctuations across evaluations~\cite{wang2023large, li2023prd, zheng2024judging, zeng2023evaluating}.
    
\paragraph{Reference-Guided Grading} Providing reference answers is crucial for evaluating models in tasks with objective human preferences, such as mathematics, translation, etc.~\cite{wang2023large, wang2023pandalm, kim2023prometheus} However, this method requires high-quality annotations for reference answers.


While this method is currently the most prevalent and boasts the highest human consistency, it suffers from scalability issues as the number of models to evaluate increases, resulting in quadratic growth in possible pairs and consequently escalating computational costs ~\cite{zeng2023evaluating}.

\subsubsection{LLM-based Evaluation Models}
The most-preferred model as an LLM-based evaluator is the priority models including GPT-4, which may be motivated by the fact that these models are typically trained using RLHF to align with human preferences and have demonstrated strong human consistency.~\cite{bai2022training}
Intuitively, evaluators should be capable of distinguishing between good and bad responses ~\cite{liu2303g, liu2303g, kocmi2023large, chiang2023can, wang2023large, li2023prd}. Utilizing state-of-the-art models in this way leads to outstanding performance and broad generalizability. However, drawbacks include high costs and the potential irreproducibility\. In response, recent research has shifted towards fine-tuning smaller, open-source LLMs for evaluation purposes, aiming to achieve performance close to GPT-4. These models are primarily created by meticulously constructing high-quality evaluation data and fine-tuning open-source models. Compared with employing priority models, Fine-tuning small models enhances the model's evaluation capability in certain aspects, mitigating the potential of bias (i.e. position bias) and significantly reducing costs.
However, experiment results indicate that while the fine-tuned evaluation models achieve superior accuracy on their respective in-domain test sets, they still exhibit limitations, including a lack of generalization, overfitting on specific evaluation strategies, and a bias towards surface quality.


\subsubsection{Limitations}
The LLM-based evaluator has been found to exhibit certain biases and limitations. ~\cite{wang2023large} note that LLM-based evaluator inevitably has position bias (i.e., when using GPT-4 for pairwise comparison, the evaluation results can be easily hacked by altering the order in which candidate answers appear in context).
Another bias of LLM-based evaluator is that the LLM-based evaluator exhibits a tendency to prefer more verbose outputs~\cite{zheng2024judging}, shows a predisposition towards outputs generated by models similar to itself ~\cite{gudibande2023false, zheng2024judging}, and displays a limited capability in evaluating subjects like mathematics, reasoning, and other areas that still pose challenges for LLMs~\cite{zeng2023evaluating}.
To systematically quantify the performance of LLM-based Evaluators, several works have introduced meta-evaluation benchmarks. FairEval~\cite{wang2023large}, MT-Bench~\cite{zheng2024judging}, and LLMEval~\cite{zhang2023wider} assess whether LLM-based evaluators demonstrate high agreement with humans by utilizing manually annotated preference datasets. ~\cite{zeng2023evaluating} proposed a meta-evaluation benchmark called LLMBar, which includes an Adversarial set.
Notably, all models, including GPT-4, struggled on the adversarial set without the use of additional strategies.

\section{Future Directions}

\paragraph{Better quality and more diverse preference data.} In the preference learning scenario, the final performance of the model to a large extent depends on the quality and diversity of the preference data~\citep{guo2024direct, song-etal-2024-scaling}. Therefore, further research can be conducted on this domain. For example, synthetic data techniques can be utilized to ensure prompt quality~\citep{yuan2024self, pace2024westofn}. Besides, advanced sampling techniques may be explored to enhance the sampling diversity and quality of the model response~\citep{xie2024montecarlotreesearch}.

\paragraph{Reliable feedback and scalable oversight.} The optimization objective of preference learning comes from the feedback, and thus reliable feedback plays an important role. Some reliable feedback such as code compiler~\citep{shen2023pangucoder2, grubisic2024compilergeneratedfeedbacklarge} or proof assistant~\citep{xin2024deepseekproverv15harnessingproofassistant} is explored, but they are limited to code or math domain. It would be valuable if we could extend them into more general domains. In addition, more research is required in cases where humans cannot provide reliable feedback anymore to enable scalable oversight for the next-generation super-intelligence, such as recursive reward modeling~\citep{yuan2024self}, or weak-to-strong technique~\citep{burns2023weaktostronggeneralizationelicitingstrong,chen2024self}.

\paragraph{Advanced algorithm for preference learning.} Data and feedback determine the upper bound of the model performance, and a good training algorithm can help us approach this upper bound as much as possible. In the future, better training algorithms should strive to meet the following requirements: (1) better approach the performance upper bound; (2) more robust to the provided data and feedback~\citep{liang2024roporobustpreferenceoptimization, ramesh2024group}; (3) higher training efficiency and therefore can be scaled up~\citep{li2023remax,meng2024simpo}. In fact, there are already many optimized variants of PPO and DPO for preference learning. However, the performance of these algorithms may be inconsistent across different models and task settings~\citep{saeidi2024insightsalignmentevaluatingdpo}. Finding the most effective variant from a theoretical perspective is also a very practical topic, which we leave to our future work.

\paragraph{More comprehensive evaluation for LLM.} The existing evaluation datasets are not comprehensive enough to assess the capabilities of models, and the form of the questions is also relatively homogeneous (e.g., multiple-choice questions). Although more and more open-ended generation evaluation benchmarks are proposed, factors such as evaluation bias~\citep{wang2023large} and the cost of evaluation~\citep{biderman2023pythiasuiteanalyzinglarge} still trouble us. We need more comprehensive, reliable, and diverse evaluation methods and benchmarks, which are complementary to the development and progress of large language models.

\section{Conclusion}
 In this survey, we decompose the strategies of preference learning into several modules: Model, Data, Feedback, and Algorithm. By distinguishing different strategies according to their variants, we build a unified view of preference learning strategies and establish connections among them. We believe that although the core objectives of these alignment algorithms are essentially similar, their performance can vary significantly across different application scenarios. We leave the exploration of which variants perform better in specific contexts as our future work. Finally, we hope this survey provides researchers with a further understanding of preference learning and thereby inspires further research in this field.

\printbibliography[heading=bibintoc]

\appendix
\newpage
\section{The discussion about Online vs. Offline}

Unlike traditional reinforcement learning, discussions about \textit{Online} and \textit{Offline} in preference learning of large language models lack a unified definition. Essentially, the distinction between online and offline lies in whether feedback is obtained in real-time from the environment, that is, whether the policy model interacts with the environment in real-time. For preference data $(x, y, r)$, if the feedback $r$ is obtained in real-time from the environment, it is considered online. For instance, in RLHF, the feedback of on-policy data is obtained in real time from the reward model during the model training. 

However, some studies~\citep{alignment_guidebook} delve deeper into whether a fixed reward model can genuinely represent the environment. These studies argue that a static reward model cannot consistently reflect the environment, thereby classifying RLHF with a fixed reward model as offline. In this review, for the sake of simplicity and clarity, we do not dwell on this distinction too much; instead, we define the output of the reward model as a form of signal from the environment. Consequently, in this context, methods like RLHF are considered online and DPO is offline.

\section{The discussion about ReMAX and GRPO}
In this chapter, we provide a detailed clarification of our discussion regarding the classification of the ReMAX~\citep{li2023remax} and GRPO~\citep{shao2024deepseekmath} algorithms. Both of these approaches aim to forego the critic model of PPO. From the perspective of the gradient coefficient, ReMax requires an additional reward from a sequence decoded greedily to serve as a baseline, while GRPO estimates the baseline from group scores. This makes ReMax a pairwise method and GRPO a listwise method.

However, from the standpoint of loss calculation, once both algorithms compute the reward or advantage, they utilize samples to optimize the model in a pointwise manner. This is because, unlike algorithms like DPO that use two or a group of samples to compute the loss itself, they focus solely on calculating the gradient coefficient. For the sake of fluency and readability in the main text, we did not include these considerations and simply categorized ReMAX as pointwise and GRPO as listwise.

\end{document}